\documentclass[letterpaper, 10 pt, journal, twoside]{ieeetran}

\usepackage{graphicx,dblfloatfix}
\usepackage{amsmath}
\usepackage{amssymb}
\usepackage{subcaption}
\usepackage{array}
\usepackage{flushend}
\usepackage{cite}
\usepackage{hyperref}
\usepackage{float}
\usepackage{mathtools}

\hypersetup{
    colorlinks=true,
    linkcolor=black,
    urlcolor=black,
    citecolor=black
}

\usepackage{amsmath}
\usepackage{amssymb}
\usepackage{accents}
\usepackage{bm}
\usepackage{xifthen}
\usepackage{color}
\usepackage{fancyvrb}
\usepackage{graphicx,scalerel}

\newcommand{\ba}{\begin{eqnarray}}
\newcommand{\ea}{\end{eqnarray}}

\newcommand{\unit}[1]{\mbox{$\mathrm{\,#1}$}}

\renewcommand{\deg}{\mbox{${}^\circ$}}

\newcommand{\presup}[1]{\,{}^{\scriptscriptstyle #1}\!}

\newcommand{\pose}[1][_NONE]{\ifthenelse{\equal{#1}{_NONE}}{}{\presup{#1}}{\mathbf{\xi}}}
\newcommand{\estpose}[1][_NONE]{\ifthenelse{\equal{#1}{_NONE}}{}{\presup{#1}}{\hat{\mathbf{\xi}}}}
\newcommand{\hpose}[1][_NONE]{\ifthenelse{\equal{#1}{_NONE}}{}{\presup{#1}}{\hat{\mathbf{\xi}}}}
\newcommand{\posedot}[1][_NONE]{\ifthenelse{\equal{#1}{_NONE}}{}{\presup{#1}}{\mathbf{\nu}}}
\newcommand{\q}[1][_NONE]{\ifthenelse{\equal{#1}{_NONE}}{}{\presup{#1}}{\mathring{q}}}
\newcommand{\uquat}[2][_NONE]{\ifthenelse{\equal{#1}{_NONE}}{}{\presup{#1}}{\mathring{#2}}}
\newcommand{\duquat}[2][_NONE]{\ifthenelse{\equal{#1}{_NONE}}{}{\presup{#1}}{\dot{\mathring{#2}}}}
\newcommand{\quat}[2][_NONE]{\ifthenelse{\equal{#1}{_NONE}}{}{\presup{#1}}{\breve{#2}}}
\DeclareMathAlphabet{\mathitbf}{OML}{cmm}{b}{it}
\newcommand{\twist}[2][_NONE]{\ifthenelse{\equal{#1}{_NONE}}{}{\presup{#1}}{\mathcal{S}}}
\renewcommand{\vec}[2][_NONE]{\ifthenelse{\equal{#1}{_NONE}}{}{\presup{#1}}{\boldsymbol #2}}
\newcommand{\hvec}[2][_NONE]{\ifthenelse{\equal{#1}{_NONE}}{}{\presup{#1}}{\tilde{\vec{#2}}}}
\newcommand{\evec}[2][_NONE]{\ifthenelse{\equal{#1}{_NONE}}{}{\presup{#1}}{\hat{\vec{#2}}}}
\newcommand{\bvec}[2][_NONE]{\ifthenelse{\equal{#1}{_NONE}}{}{\presup{#1}}{\bar{\vec{#2}}}}
\newcommand{\dhvec}[2][_NONE]{\ifthenelse{\equal{#1}{_NONE}}{}{\presup{#1}}{\dot{\tilde{\vec{#2}}}}}
\newcommand{\dvec}[2][_NONE]{\ifthenelse{\equal{#1}{_NONE}}{}{\presup{#1}}{\dot{\vec{#2}}}}
\newcommand{\ddvec}[2][_NONE]{\ifthenelse{\equal{#1}{_NONE}}{}{\presup{#1}}{\ddot{\vec{#2}}}}

\newcommand{\mat}[2][_NONE]{\ifthenelse{\equal{#1}{_NONE}}{}{\presup{#1}\,}{{\mathbf #2}}}
\newcommand{\dmat}[2][_NONE]{\ifthenelse{\equal{#1}{_NONE}}{}{\presup{#1}\,}{{\dot{\mathbf #2}}}}
\newcommand{\emat}[2][_NONE]{\ifthenelse{\equal{#1}{_NONE}}{}{\presup{#1}\,}{\hat{\mathbf#2}}}
\newcommand{\matfn}[3][_NONE]{\ifthenelse{\equal{#1}{_NONE}}{}{\presup{#1}}{{\mat{#2}}\left(#3\right)}}
\newcommand{\Rt}[2][_NONE]{\ifthenelse{\equal{#1}{_NONE}}{}{\presup{#1}}{{\bf R}\left(#2\right)}}
\newcommand{\cframe}[1]{\ensuremath{\{\mathrm{#1}\}}}

\newcommand{\point}[2][_NONE]{\ifthenelse{\equal{#1}{_NONE}}{}{\ensuremath{\presup{#1}}}{\ensuremath{\mathrm{#2}}}}

\newfont{\School}{pncr}
\newfont{\eightTR}{pncr at 8pt}

\fvset{formatcom={\color{blue}\baselineskip=-\maxdimen\lineskip=0pt},xleftmargin=4mm,commentchar=!}
\DefineVerbatimEnvironment{Code}{Verbatim}{}
\DefineVerbatimEnvironment{CodeSmall}{Verbatim}{fontsize=\scriptsize,xleftmargin=1mm,commentchar=!}
\DefineVerbatimEnvironment{CodeNum}{Verbatim}{numbers=left,numbersep=4pt,fontsize=\footnotesize,xleftmargin=4mm}
\newcommand{\func}[2][_NONE]{\ifthenelse{\equal{#1}{_NONE}}{\index{code}{#2}}{\index{code}{#1}}\ifthenelse{\boolean{draft}}{{\color{green}\Verb+#2+}}{\Verb+#2+}}
\newcommand{\methodb}[2]{\index{code}{#1@\textbf{#1}!.#2}\ifthenelse{\boolean{draft}}{{\color{magenta}\Verb+#1.#2+}}{\Verb+#1.#2+}}
\newcommand{\method}[2]{\index{code}{#1@\textbf{#1}!.#2}\ifthenelse{\boolean{draft}}{{\color{magenta}\Verb+#2+}}{\Verb+#2+}}
\newcommand{\class}[1]{\index{code}{#1@\textbf{#1}}\ifthenelse{\boolean{draft}}{{\color{cyan}\Verb+#1+}}{\Verb+#1+}}
\newcommand{\property}[1]{\index{property}{#1}\ifthenelse{\boolean{draft}}{{\color{cyan}\Verb+#1+}}{\Verb+#1+}}

\newcommand{\SE}[1]{\ensuremath{\mathrm{{\bf SE}(#1)}}}

\usepackage{xparse}
\NewDocumentCommand{\ovec}{ o m o }{%
  \IfValueT{#1}{\presup{#1}}%
  \vec{#2}%
  \IfValueT{#3}{_{#3}}
  ^{\scalebox{0.6}{\#}}%
  }
\NewDocumentCommand{\omat}{ o m o }{%
  \IfValueT{#1}{\presup{#1}}%
  \mat{#2}%
  \IfValueT{#3}{_{#3}}
  ^{\scalebox{0.6}{\#}}%
  }
\NewDocumentCommand{\obspose}{ o o }{%
  \IfValueT{#1}{\presup{#1}}%
  \mathbf{\xi}%
  \IfValueT{#2}{_{#2}}
  ^{\scalebox{0.6}{\#}}%
  }

\begin{document}

\title{A Holistic Approach to Reactive Mobile Manipulation}

\author{Jesse Haviland, Niko Sünderhauf,  Peter Corke
\thanks{Manuscript received: September, 9, 2021; Revised December, 5, 2021; Accepted January, 10, 2022.}
\thanks{This paper was recommended for publication by Editor M. Vincze upon evaluation of the Associate Editor and Reviewers' comments. This research is supported by the Queensland University of Technology (QUT) Centre for Robotics.}
\thanks{Jesse Haviland, Niko Sünderhauf, and Peter Corke are with the QUT Centre for Robotics (QCR), Brisbane, Australia
{\tt\small \{j.haviland, niko.suenderhauf, peter.corke\}@qut.edu.au}.
}
\thanks{Digital Object Identifier (DOI): 10.1109/LRA.2022.3146554}%
}

\markboth{IEEE Robotics and Automation Letters. Preprint Version. Accepted January, 2022}
{Haviland \MakeLowercase{\textit{et al.}}: A Holistic Approach to Reactive Mobile Manipulation} 

\maketitle

\begin{abstract}

We present the design and implementation of a taskable reactive mobile manipulation system. In contrary to related work, we treat the arm and base degrees of freedom as a holistic structure which greatly improves the speed and fluidity of the resulting motion. At the core of this approach is a robust and reactive motion controller which can achieve a desired end-effector pose, while avoiding joint position and velocity limits, and ensuring the mobile manipulator is manoeuvrable throughout the trajectory. This can support sensor-based behaviours such as closed-loop visual grasping. As no planning is involved in our approach, the robot is never stationary \emph{thinking} about what to do next. We show the versatility of our holistic motion controller by implementing a pick and place system using behaviour trees and demonstrate this task on a 9-degree-of-freedom mobile manipulator. Additionally, we provide an open-source implementation of our motion controller for both non-holonomic and omnidirectional mobile manipulators available at \href{https://jhavl.github.io/holistic}{jhavl.github.io/holistic}.

\end{abstract}

\begin{IEEEkeywords}
    Mobile Manipulation, Motion Control, Kinematics, and Reactive Control
\end{IEEEkeywords}


\section{Introduction}
\IEEEPARstart{P}{rogramming} a robot to perform a task is a long-standing challenge in robotics and most systems employ some combination of planning and reactivity. Planning takes a current and desired world state and outputs a robot trajectory to effect that change. Solutions are optimal with respect to some measure but the plan is executed open-loop and its success is therefore critically reliant on the provided initial world state and the accuracy of the robot in following the plan. Planning systems today are extremely capable and can handle problems with many degrees of freedom and constraints, but the planning time can be a significant fraction of a second or more.

In contrast, reactive systems ``close the loop'' on sensory information.  Such systems are highly responsive and robust to changes in their environment but typically do not handle complex constraints and the resulting behaviours are typically simplistic and not suitable for most real-world tasks.

The planning-reactivity dichotomy was identified long ago \cite{nasrem} and most robot systems today combine planning and reactive subsystems. For example, the global and local planner architecture is commonly used, and the ``local planner" provides the system's reactivity to the real world.

In this letter, we extend our earlier work on reactive control for robot manipulator arms \cite{neo} to a mobile manipulator that includes a non-holonomic base.  This provides the ability to command end-effector velocity in an infinitely large workspace with motion optimally partitioned between arm and base in a way that maximises arm dexterity and allows graceful and fluid motion of the whole mobile manipulator. We achieve holistic motion by considering all degrees-of-freedom simultaneously within a quadratic program.

However, there remains a significant gap between what a reactive behaviour can achieve and what is necessary to complete a useful task. A task is typically a sequence of behaviours which admits the need for detecting when a behaviour has completed, switching between behaviours according to world state and the goal, and error recovery.  

\begin{figure}[t]
    \centering
    \includegraphics[height=4.9cm]{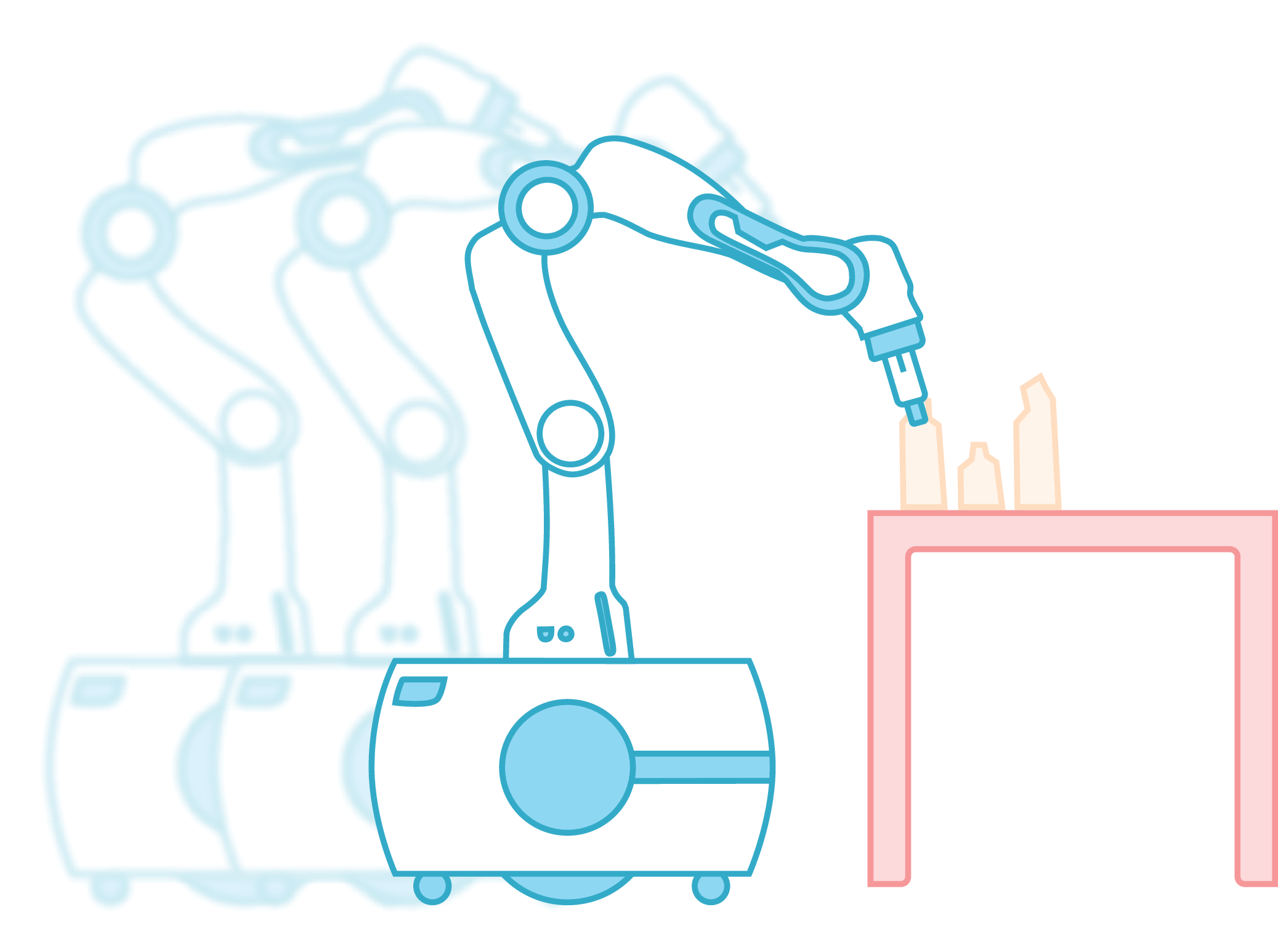}
    \caption{
        Our reactive motion controller treats the mobile manipulator's base and arm degrees of freedom as a holistic system. The resulting system can perform closed-loop visual grasping in dynamic environments while avoiding joint position and velocity limits and ensuring the mobile manipulator remains manoeuvrable throughout the motion.
    }
    \vspace{-10pt}
\end{figure}

To make our reactive system taskable we combine it with a behaviour tree. Compared to alternatives such as state machines, behaviour trees are easier to express and have inherent error recovery capability. The behaviour tree orchestrates the sensor-based behaviours of the reactive subsystem to achieve useful tasks. The behaviour tree can be considered as a form of scripting language, and still requires human ingenuity to create but could, in the future, be synthesised by a task-level planner. The result is a holistic system that can perform complex tasks while exhibiting fluid and continuous motion without pauses for planning.

The contributions of this letter are:
\begin{enumerate}
    \item a novel method of incorporating a non-holonomic mobile manipulation platform into a holistic reactive motion control framework with example code available at \href{https://jhavl.github.io/holistic}{jhavl.github.io/holistic}
    \item complete integration of a closed-loop mobile grasper using behaviour trees
    \item reactive control and validation on a 9 degree of freedom non-holonomic mobile manipulation platform with a comparison to prior work.
\end{enumerate}

\section{Related Work}\label{sec:related_work}
Many mobile manipulation systems have been designed in recent years. Most prominently, the work in \cite{mm0} used visual, tactile and proprioceptive sensor data to accomplish pick and place tasks using the PR2 robot. Graspable objects and support surfaces are segmented from 3D data, known objects have meshes fitted, and point clouds represent unknown objects. Grasp poses are achieved using an OMPL motion planner \cite{ompl} which generates motion for the 7 degree-of-freedom arm to follow. The approach in \cite{mm1} extends this to tightly couple visual and tactile sensors to enable grasp failure detection. The tactile feedback enables some local reactivity in grasping to account for sensor error or small grasp-target perturbances. Similar approaches are used on the Intel HERB platform in \cite{mm7, mm9}. Additionally, \cite{mm8} uses a similar approach on the ARMAR-III robot for dual-arm grasping. In each of these works, the base of the robot is moved into position \emph{before} actuating the arm to complete a grasp. While this simplifies the problem, the task is completed in a slow, and discontinuous manner where the motion is not \emph{graceful}. Graceful motion is defined as being safe, comfortable, fast, and intuitive \cite{grace}. Specifically, the velocity must be smooth and consistent -- not stop-start or unnatural.

Each of the above-mentioned works use motion planning to find an optimal trajectory subject to the supplied criteria. Motion planners can offer collision avoidance, joint limit avoidance, among many useful features. However, the associated computation time limits them to open-loop control. In a benchmark from \cite{trajopt}, motion planners from the OMPL library range from 0.61\unit{s} to 1.3\unit{s} to compute a trajectory on a 7 degree-of-freedom arm, and between 18.7\unit{s} to 20.3\unit{s} to compute a trajectory on an 18 degree-of-freedom PR2 robot. This look-then-move paradigm is slow and not robust to environmental change and sensor or motion error. 

The NEO controller in \cite{neo} exploits the differential kinematics of a table-top manipulator to create a motion controller which can avoid both static and dynamic obstacles while also avoiding the joint position and velocity limits of the manipulator. The controller is purely reactive and is an extension of the classical resolved-rate motion control \cite{rrmc} which has been re-framed as a quadratic programming optimisation problem (QP). NEO's performance was shown to be comparable to motion planners for a reaching task with obstacles -- while taking 9.8\unit{ms} to compute each step. An alternate approach to reactive motion control is through potential fields where the robot is repelled from joint limits and obstacles while being attracted towards the goal \cite{re0, pot} and has been implemented on a mobile manipulator \cite{re2}. However, the potential field approach performs poorly compared to NEO \cite{neo} and becomes intractable for complex and cluttered environments. In \cite{re1}, reactive whole-body control was achieved on a holonomic humanoid platform using a velocity controller on the mobile base and force control for the manipulators. The low latency of reactive motion controllers allows for greater versatility in the operation of the robot than planning based approaches, and makes tasks such as closed-loop visual grasping possible.

Reactive control of a non-holonomic mobile base is challenging since it cannot achieve an arbitrary task-space velocity through a time-invariant state-feedback controller \cite{nonhom}. This can be overcome with a switched control algorithm \cite{switch} or a dynamics-based approach \cite{mm0}. The most prominent method of controlling a mobile base is with non-linear optimisation based motion planning \cite{latombe}.

Combining and incorporating various controllers into a mobile manipulation system can be just as difficult as designing the individual components \cite{mm0}. An alternate control mechanism, Riemannian Motion Policies, fuses the outputs from different motion controllers to actuate the robot \cite{rmp}. However, this concept is confined to motion generation and lacks the ability to coordinate perception, navigation and actuation. A common approach is to use finite-state machines (FSM) or similar architectures \cite{mm1}.  In recent times, behaviour trees have grown in popularity, and have been shown to be very useful for robotics where complex tasks need to be completed in a robust, reliable, and reactive manner \cite{bt1}. Behaviour trees provide an alternative to FSMs and have several advantages such as readability, maintainability, and scalability, while also providing a greater degree of expressiveness \cite{bt0}. The reactive nature of behaviour trees provides a better match for reactive motion controllers than FSMs as robot targets and state information can be updated quickly and easily based on visual/robot feedback without requiring very complicated and unreadable state flow.
Additionally, these benefits extend to reactive error recovery which can be implemented simply and based on visual/robot feedback.

\section{Modelling a Mobile Manipulator}\label{sec:mobile}
Mobile manipulators comprise a mobile base with one or more manipulators where the base is either omnidirectional or non-holonomic.

\subsection{Two-Wheel Differential Drive Mobile Manipulators}

The pose of the robot's end-effector is
\begin{equation}
\mat[0]{T}_e =\mat[0]{T}_b(x, \ y, \ \theta) \mat[b]{T}_a \mat[a]{T}_e(\kappa_a, \vec{q_a}) \label{eq:fkine}
\end{equation}
where \cframe{0} is the world reference frame, $\mat[0]{T}_b$ is the relative pose of the mobile base \cframe{b}, $\mat[b]{T}_a$ is a constant relative pose from the mobile base frame to the base of the manipulator arm \cframe{a}, $\mat[a]{T}_e$ is the forward kinematics of the arm where the end-effector frame is \cframe{e}, $\kappa_a$ are the arm kinematic parameters and $\vec{q_a}$ are the arm's joint coordinates.

\begin{figure*}[t!]
    \vspace{6pt}
    \centering
    \includegraphics[width=16.8cm,height=5cm]{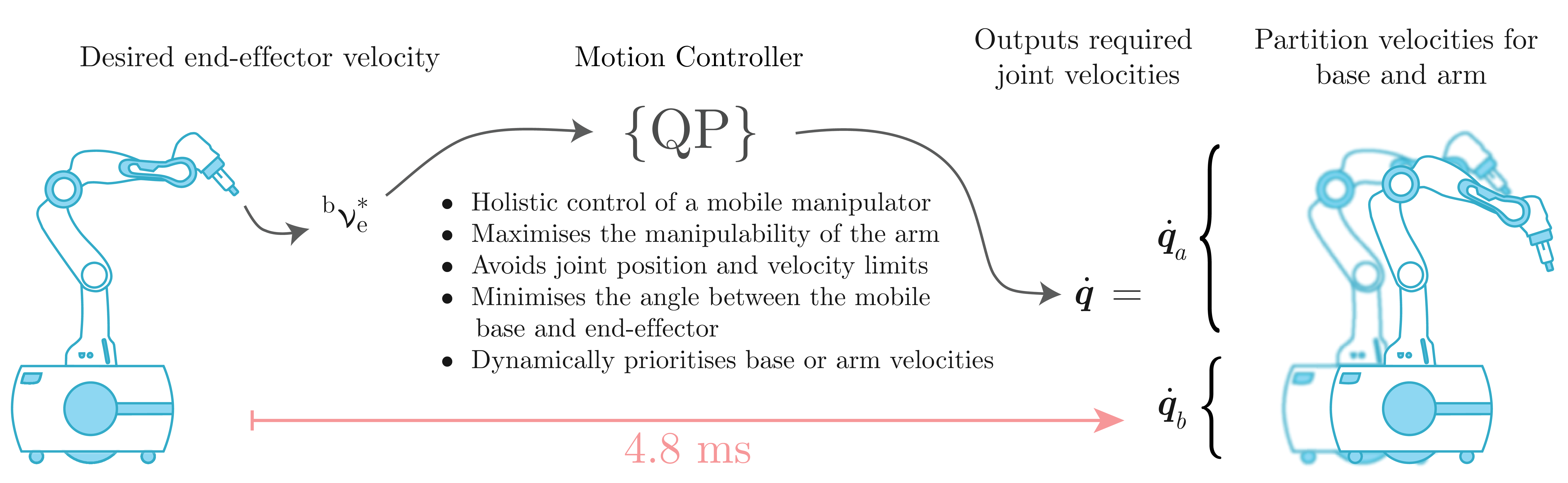}
    \caption{
        The desired end-effector spatial velocity ${^b\vec{\nu}^*_e}$ is taken as the input to our motion controller, which is expressed as a quadratic programme (QP). The motion controller minimises defined costs and incorporates constraints to ensure reliable, fast, and graceful control of the robot. The QP outputs joint velocities $\dvec{q}$ to be actuated by the robot. We partition these joint velocities into the base virtual joint velocities $\dvec{q}_b$ and arm joint velocities $\dvec{q}_a$. The arm joint velocities are applied directly to the arm, while the base virtual joint velocities are transformed and applied to mobile base's wheels as described in Section \ref{sec:mobile}.
    }
    \label{fig:whole}
\end{figure*}

Our approach is based on differential kinematics and we wish to determine a relationship between base velocity and joint angle rates $\dvec{q}_a$, and the end-effector spatial velocity $\posedot_e$. To facilitate this we introduce an infinitesimal transform into the kinematic chain (\ref{eq:fkine}) to represent base instantaneous motion
\begin{equation}
    \mat[0]{T}_e =\mat[0]{T}_b \underbrace{\mat[b]{T}_{b'}(\delta_\theta, \delta_d)}_{\mbox{\tiny base instantaneous}} \mat[b']{T}_a \mat[a]{T}_e(\kappa_a, \ \vec{q_a}) \label{eq:diffkine}
\end{equation}
where $\delta_\theta$ and $\delta_d$ are respectively infinitesimal rotation and forward translation of the non-holonomic base.  We could consider this as a simple revolute-prismatic robot with virtual joint coordinates $\vec{q_b} \in \mathbb{R}^2, \|\vec{q_b} \| \to 0$ and $\dot{q}_{b,0} = \dot{\delta}_\theta$ and $\dot{q}_{b,1} = \dot{\delta}_d$. The resulting expression is now similar to the forward kinematics of a serial-link manipulator and using the approach of \cite{ets} we can write the differential kinematics as
\begin{align} 
    \posedot[0]_e =
    \mat[0]{J}_e(x, y, \theta, \vec{q}) \
    \begin{pmatrix} \dvec{q}_b \\ \dvec{q}_a \end{pmatrix}
\end{align}
which maps the velocity of all axes of the mobile manipulation system to the end-effector spatial velocity where $\vec{q} = (\vec{0}_2, \ \vec{q}_a)^\top \in \mathbb{R}^n$ and $n$ is the number of virtual and real joints. The holistic manipulator Jacobian $\mat[0]{J}_e(\cdot)$ is expressed in the world frame. The velocity of the virtual base joints are related to the wheel angular velocities by
\begin{align} 
    \dot{q}_{b,0} = \tfrac{R}{W}(\varpi_R-\varpi_L), \,\dot{q}_{b,1} = \tfrac{R}{2}(\varpi_R + \varpi_L)  \label{eq:dd1}
\end{align}
where $R$ is the wheel radius, $\varpi_R$, $\varpi_L$ are the robot's right and left wheel velocities respectively, and $W$ is the distance between the two wheels. 

For some control behaviours, it is useful to express velocity in different frames. For example, in the base frame $\posedot[b]_e$ or end-effector frame $\posedot[e]_e$ -- the corresponding Jacobians ${^b}\mat{J}_e(\vec{q})$ and ${^e}\mat{J}_e(\vec{q})$ can be similarly derived.

Using the holistic manipulator Jacobian, we can construct a resolved-rate motion controller 
\begin{equation}  \label{eq:rrmc1}
    \dvec{q}^*(t) = {^b\mat{J}_e}(\vec{q})^{+} \ {^b\vec{\nu}^*_e}(t)
\end{equation}
where $(\cdot)^+$ denotes the Moore-Penrose pseudoinverse, $\dvec{q}^*$ is the joint velocities required for the mobile manipulator's end-effector to experience a desired spatial velocity and ${^b\vec{\nu}^*_e} = (v_x, v_y, v_z, \omega_x, \omega_y, \omega_z )^\top$. This controller represents the most basic form of reactive and holistic control of a mobile manipulator.

The demanded joint velocity $\dvec{q}^*$ is partitioned into the base virtual joint velocities $\dvec{q}^*_b$ which are transformed to the left and right wheel velocities using the unicycle kinematic model using (\ref{eq:dd1}), and $\dvec{q}^*_a$ corresponds to the arm's joint velocities.

\subsection{Omnidirectional Mobile Manipulators}

Omnidirectional robots can instantaneously translate and rotate in the plane and using the virtual joint approach from above
we define $\vec{q}_b = (\delta_x, \delta_y, \delta_\theta) \in \mathbb{R}^3$.

\section{Mobile Manipulator Motion Controller}\label{sec:cont}
We extend the NEO motion controller from \cite{neo} which was designed for table-top manipulators to enable holistic control of a mobile manipulator. We provide a high level overview of our approach in Figure \ref{fig:whole}. This controller outputs joint velocities $\dvec{q}$ that achieve a desired end-effector velocity ${^b\vec{\nu}^*_e}$ minus a slack vector (intentional error to the desired velocity) $\vec{\delta}$ where the slack component is used to allow the controller to better minimise costs and meet certain constraints.
This controller is expressed as a QP
\begin{align} \label{eq:qp0}
    \min_x \quad f_o(\vec{x}) &= \frac{1}{2} \vec{x}^\top \mathcal{Q} \vec{x}+ \mathcal{C}^\top \vec{x}, \\ 
    \mbox{subject to} \quad \mathcal{J} \vec{x} &= {^b\vec{\nu}}_e, \nonumber \\
	\mathcal{A} \vec{x} &\leq \mathcal{B} \nonumber \\
	\mathcal{X}^- &\leq \vec{x} \leq \mathcal{X}^+ \nonumber
\end{align}
where 
$\mat{x}=(\vec{q}, \ \vec{\delta})^\top$ is the decision variable,
$\mathcal{Q}$ incorporates joint velocity and slack costs,
$\mathcal{C}$ incorporates manipulability maximisation and auxiliary performance tasks,
$\mathcal{J}$ incorporates an augmented manipulator Jacobian,
$\mathcal{A}$ and $\mathcal{B}$ implement joint position limit avoidance, and
$\mathcal{X}^{+, -}$ limits maximum and minimum values of the decision variable.
To lift the capabilities of NEO \cite{neo} from stationary to mobile manipulators, we make significant changes to formulation of the QP which we detail below.

\subsection{Velocity Control}

The end-effector has a spatial velocity
\begin{equation}
    {^b\vec{\nu}}_e(t) = {^b\vec{\nu}}_e^*(t) - \vec{\delta}(t)  \in \mathbb{R}^6
\end{equation}
where ${^b\vec{\nu}}_e^*(t) \in \mathbb{R}^6$ is the desired end-effector spatial velocity. $\vec{\delta}(t)\in \mathbb{R}^6$ is the slack velocity which allows for the end-effector to stray from the desired motion in order to satisfy the inequality constraints. This motion is created from the equality constraint in (\ref{eq:qp0}) with
\begin{align}
	\mathcal{J} &=
	\begin{pmatrix}
		{^b\mat{J}}_e(\vec{q}) & \mat{1}_{6 \times 6}
	\end{pmatrix} \in \mathbb{R}^{6 \times (n+6)} \\
	\vec{x} &= 
	\begin{pmatrix}
		\dvec{q} \\ \vec{\delta}
    \end{pmatrix} \in \mathbb{R}^{(n+6)}
\end{align}
where $\mat{1}$ is the identity matrix.

\subsection{The Objective Function}

The optimisation will minimise $\vec{x}$ in the objective function subject to the equality and inequality constraints. The objective function in (\ref{eq:qp0}) seeks to minimise the joint velocity norm, minimise the slack norm, and maximise the manipulability of the robot according to the measure from \cite{manip}. More information on manipulability maximisation is available in \cite{neo, mmc, manip2}. The objective function is defined by
\begin{align}
	\mathcal{Q} &=  \label{eq:gain0}
	\begin{pmatrix}
		\mbox{diag}(\vec{\lambda}_q) & \mathbf{0}_{6 \times 6} \\ \mathbf{0}_{n \times n} & \mbox{diag}(\vec{\lambda}_\delta)
	\end{pmatrix} \in \mathbb{R}^{(n+6) \times (n+6)}, \\
	\mathcal{C} &=  \label{eq:gain1}
	\begin{pmatrix}
		\emat{J}_m + \vec{\epsilon} \\ \mat{0}_{6 \times 1}
    \end{pmatrix} \in \mathbb{R}^{(n + 6)}
\end{align}
where $\vec{\lambda}_q \in \mathbb{R}^n$ adjusts between minimising the joint velocity norm and maximising manipulability, $\vec{\lambda}_\delta \in \mathbb{R}^n$ adjusts the cost of adding slack, $\emat{J}_m$ is the manipulability Jacobian, and $\vec{\epsilon}$ is the base to end-effector angle.

\textit{Manipulability $\emat{J}_m$}: The manipulability metric $m(\vec{q})$ describes how easily a manipulator can achieve an arbitrary end-effector velocity. The base is capable of providing 2 (or 3 for a holonomic platform) degrees-of-freedom to contribute towards the end-effector velocity with an infinite workspace. In contrast, the arm will typically have more degrees-of-freedom but a finite workspace. To ensure the arm is capable of exploiting the extra degrees-of-freedom to provide any arbitrary end-effector velocity at a given time, we calculate the manipulability Jacobian using the arm joints
\begin{align} \label{eq:jmref}
\emat{J}_m =
		\begin{pmatrix}
        \mat{0}_b, & \mat{J}^a_m(\vec{q}_a)
		\end{pmatrix}
\end{align}
before being used in (\ref{eq:gain1}), where $\mat{0}_b$ is a zero vector, and $\mat{J}^a_m(\vec{q}_a)$ is the manipulability Jacobian corresponding to arm joints \cite{neo}.

\begin{figure}[t!]
    \vspace{6pt}
    \centering
    \includegraphics[height=5cm]{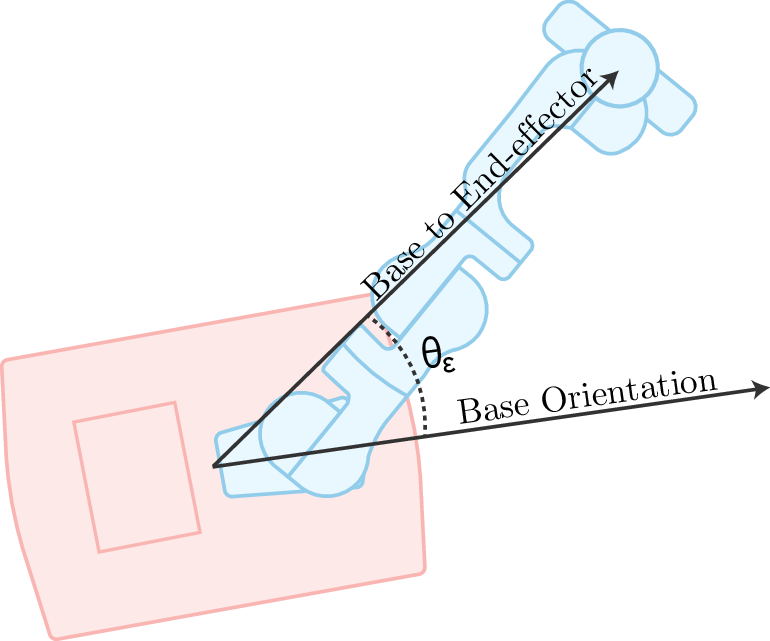}
    \caption{
        The motion controller seeks to minimise the angle $\theta_\epsilon$ between the base orientation and the end-effector which ensures that the mobile manipulator remains highly manoeuvrable.
    }
    \vspace{-10pt}
    \label{fig:orient}
\end{figure}

\textit{Base Orientation $\theta_\epsilon$}:
Figure \ref{fig:orient} shows the base to end-effector angle $\theta_\epsilon$.
When $\theta_\epsilon \simeq 0^{\circ}$ the platform has infinite reach in the forward direction but when $\theta_\epsilon \simeq \pm 90^{\circ}$, the platform's reach is limited to that of the arm. In our approach, the end-effector leads towards the goal, therefore, maximum reach is achieved when $\theta_\epsilon \simeq 0^{\circ}$.
The linear component of the cost function causes the optimiser to choose joint velocities which orient the mobile base towards the end-effector as described by Figure \ref{fig:orient}.
This cost is
\begin{align} \label{eq:kappa}
	\vec{\epsilon} =
	\begin{matrix*}[l]
		\begin{pmatrix}
			-k_\epsilon \theta_\epsilon,
			& \mathbf{0}_{n-1}
		\end{pmatrix}
	\end{matrix*}
\end{align}
where $\theta_\epsilon =
	\mbox{atan2}
	\left(
		{^b\mat{T}}_{e_{1,3}}, {^b\mat{T}}_{e_{0,3}}
	\right)$ and
$k_\epsilon$ is a gain which adjusts how aggressively the base will be re-oriented.

\textit{Slack Cost $\lambda_\delta$}: The slack cost determines the size of the slack vector in the output of the optimiser. A large value limits the slack possible and associated benefits, while a small value leads to the possibility that the slack will cancel out the desired velocity, leaving a large steady-state error. We choose $\lambda_\delta = \frac{1}{\| \vec{e} \|}$ where $\| \vec{e}\|$ is the Euclidean distance error between the end-effector and the desired end-effector pose.
This provides the optimiser the freedom to avoid operational limits and maximise manipulability at the beginning of the trajectory, while the increasing restriction ensures the end-effector continues to approach the goal. 

\textit{Joint Velocity Cost $\lambda_q$}: 
Robot arms have a finite reach and a limited joint range in which they are highly manipulable, so we choose the joint velocity cost in the QP
\begin{align} \label{eq:ka}
	\vec{\lambda}_q =
	\begin{pmatrix}
		\frac{\mat{1}}{\| \vec{e} \|}_b, & k_{a}
	\end{pmatrix}
\end{align}
where the variable gain applies to the mobile base joints, and the constant $k_a$ is applied to the arm joints. This causes the optimiser to favour the mobile-base degrees of freedom when the error $\| \vec{e} \|$ is large, and the arm degrees of freedom when the error is small.

\subsection{Joint Position Limit Avoidance}

Joint position-limit avoidance is implemented using velocity dampers. The velocity damper will restrict joint velocities of the robot to reduce or nullify the rate at which it is approaching a joint limit. Note that we do not perform joint position limit avoidance on the virtual base joints. The joint position-limit avoidance velocity damper is defined as
\begin{align}
	\mathcal{A} &= 
	\begin{pmatrix}
		\mat{1}_{n \times n + 6} \\
	\end{pmatrix} \in \mathbb{R}^{(n \times n + 6)}
\end{align}
\begin{align} \label{eq:gain2}
\mathcal{B} &= 
	\begin{pmatrix}
		\mat{0}_b \\
		\eta
		\frac{\rho_0 - \rho_s}
			{\rho_i - \rho_s} \\
		\vdots \\
		\eta
		\frac{\rho_n - \rho_s}
			{\rho_i - \rho_s}
	\end{pmatrix} \in \mathbb{R}^{n}
\end{align}
where $\rho$ is the distance to the nearest joint limit, $\rho_i$ is the influence distance in which to operate the damper, and $\rho_s$ is the minimum distance for a joint to its limit.

\subsection{Position-Based Servoing}
We use the motion controller within a position-based servoing (PBS) scheme. This controller seeks to drive the robot's end-effector from its current pose to a desired pose. PBS is formulated as
\begin{equation} \label{eq:pbs}
    {^b\vec{\nu}^*_e} = \beta \ \psi\left( ({^b\mat{T}}_e)^{-1} \ {^b\mat{T}}_{e^*} \right)
\end{equation}
where $\beta \in \mathbb{R}^+$ is a gain term, ${^b\mat{T}}_{e^*} \in \SE{3}$ is the desired end-effector pose in the robot's base frame, and $\psi(\cdot) : \mathbb{R}^{4\times 4} \mapsto  \mathbb{R}^6$ is a function which converts a homogeneous transformation matrix to a spatial displacement.

\section{Experiments} \label{sec:exp}
We first validate our motion controller by demonstrating how a mobile manipulator travels between various poses, before demonstrating our proposed architecture
performing a mobile pick and place behaviour task. 

\subsection{Platform Description}

We use our in-house developed mobile manipulation platform Frankie for the following experiments. Frankie comprises an Omron LD-60 two-wheel differential-drive base and a 7 degree-of-freedom Franka-Emika Panda manipulator. In Experiment 1 and 2 we additionally use an Omni-Frankie model in simulation which has a holonomic base (which could be realised using mecanum wheels) to exhibit holistic operation with a holonomic mobile manipulator.
The models of Frankie and Omni-Frankie are provided as part of the Robotics Toolbox for Python \cite{rtb}. For real-world experiments, we use an Intel RealSense D435 RGB-D camera attached to the end-effector for visual feedback to support visual grasping. The Omron LD-60 mobile base performs localisation and mapping on an embedded computer using an in-built LiDAR. We control the platform using the software package from \href{https://github.com/qcr/ros_omron_agv}{github.com/qcr/ros\_omron\_agv}. Frankie contains an Intel i5 NUC and an Nvidia Xavier to perform image processing.

\subsection{Experiment 1: Holistic Motion Controller Validation}

In this simulated experiment, we first show how our proposed motion controller responds to various desired end-effector poses. Starting from the same initial pose, the base and arm must move to achieve a desired end-effector pose

\renewcommand{\theenumi}{\alph{enumi}}
\begin{enumerate}
    \item 4\unit{m} in front of the robot.
    \item 4\unit{m} to the right of the robot.
    \item 4\unit{m} behind the robot.
    \item 4\unit{m} in front of the robot, where the target moves forwards and to the left for 6\unit{s} and then to the right for 6\unit{s} before becoming stationary.
\end{enumerate}

The distance 4\unit{m} was chosen as it is sufficient to require base and arm motion.

\subsection{Experiment 2: Random Reaching Task}
In this simulated random reaching task, the mobile manipulator must reach 1000 randomly generated valid and reachable end-effector poses, located within a 4\unit{m} radius of $\{0\}$. Using this methodology, we:

\begin{enumerate}
    \item compare the success rate, on both Frankie and Omni-Frankie, of our Holistic controller with separate base and arm planners (using RRTConnect algorithm from OMPL \cite{ompl}) which represents the approach from \cite{mm0}.
    \item compare the success rate of Frankie using our holistic controller while adjusting $k_\epsilon$ introduced in Section \ref{sec:cont}.
    \item compare the success rate of Frankie using our holistic controller while adjusting $\emat{J}_m$ introduced in Section \ref{sec:cont}.
\end{enumerate}

\begin{figure*}[t!] 
    \centering
    \begin{subfigure}{0.49\textwidth}
        \vspace{8pt}
        \centering
        \includegraphics[width=8.4cm]{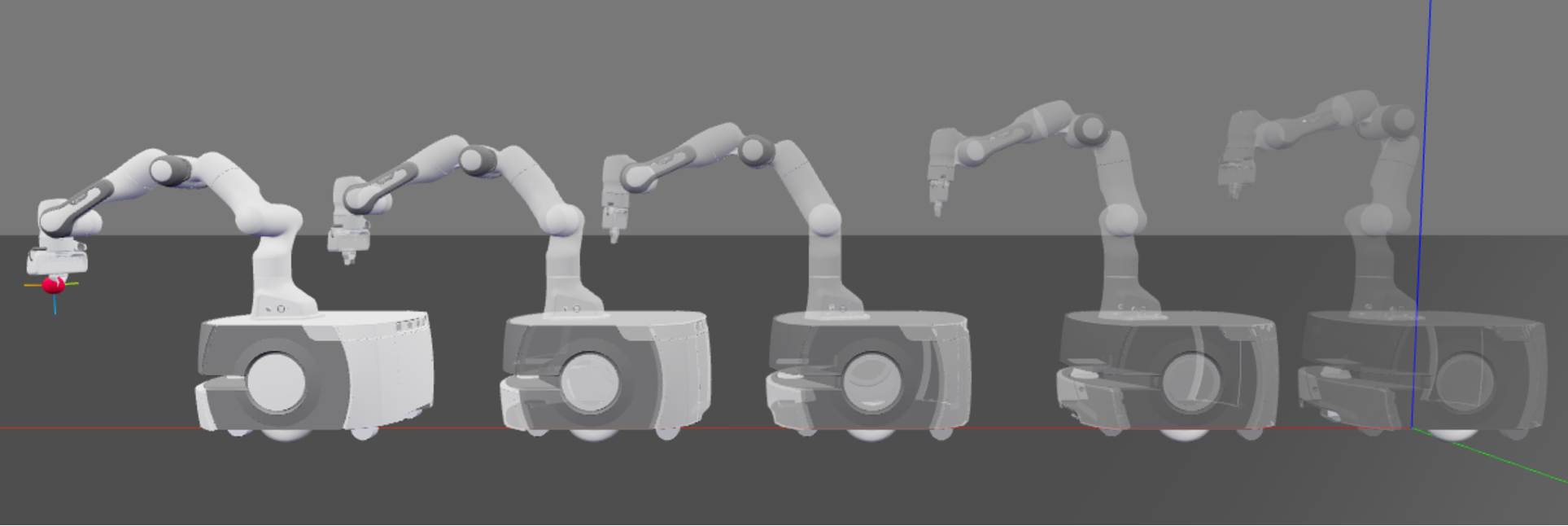}
        \caption{Experiment 1a) goal 4\unit{m} in front}
        \label{fig:p1}
    \end{subfigure}
    \hfill
    \begin{subfigure}{0.49\textwidth}
        \vspace{8pt}
        \centering
        \includegraphics[width=8.4cm]{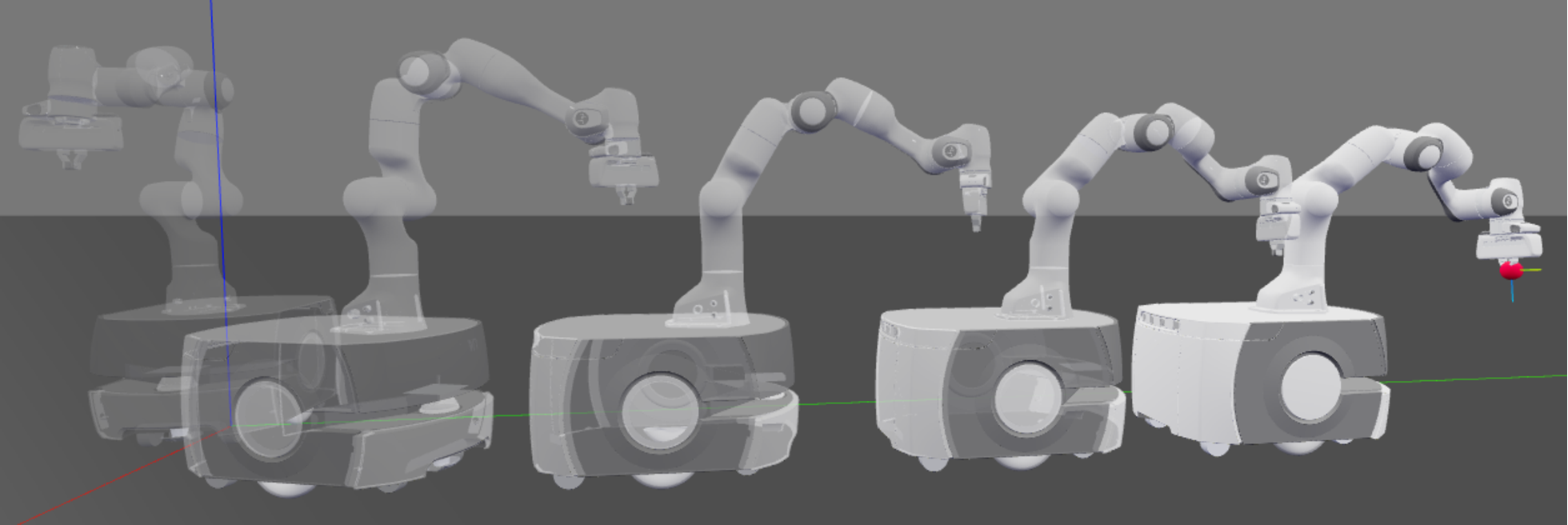}
        \caption{Experiment 1b) goal 4\unit{m} to the right}
        \label{fig:p2}
    \end{subfigure}
    
    \begin{subfigure}{0.49\textwidth}
        \vspace{8pt}
        \centering
        \includegraphics[width=8.4cm]{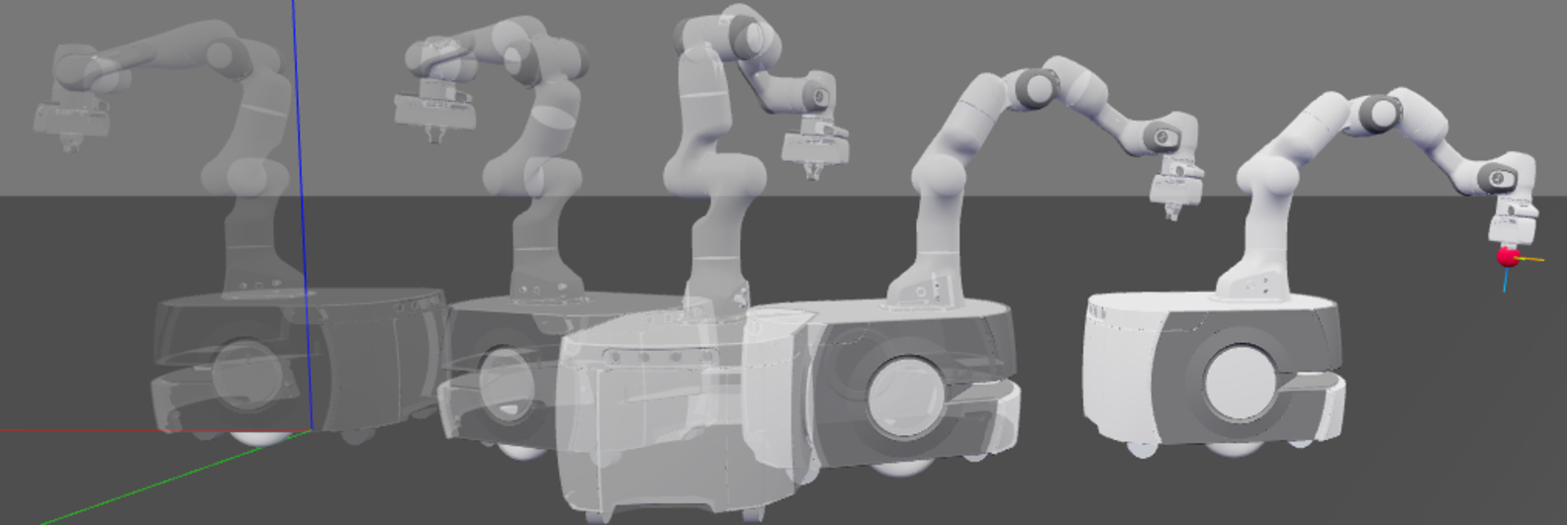}
        \caption{Experiment 1c) goal 4\unit{m} behind}
        \label{fig:p3}
    \end{subfigure}
    \hfill
    \begin{subfigure}{0.49\textwidth}
        \vspace{8pt}
        \centering
        \includegraphics[width=8.4cm]{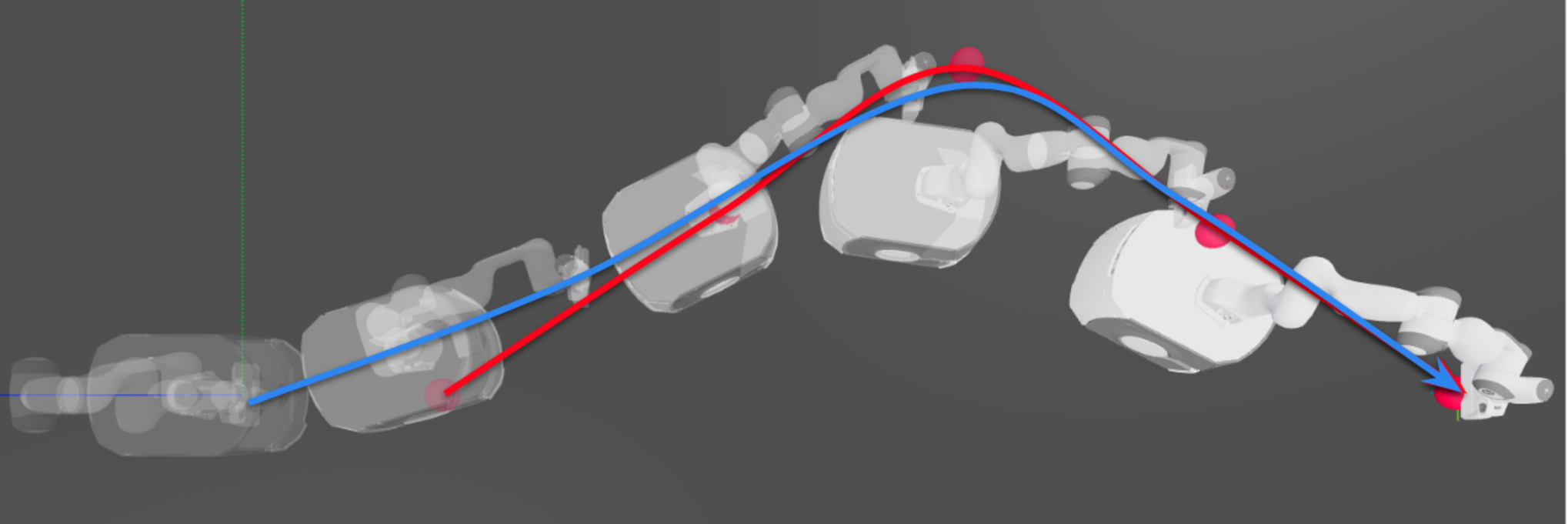}
        \caption{Experiment 1d) moving goal}
        \label{fig:move}
    \end{subfigure}

    \begin{subfigure}{0.49\textwidth}
        \vspace{4pt}
        \centering
        \includegraphics[width=8.4cm]{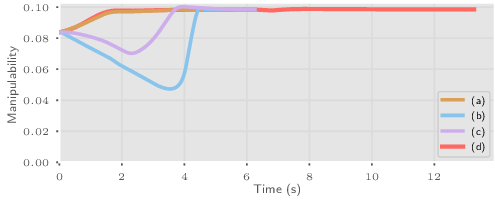}
        \vspace{-4pt}
        \caption{Manipulability of the Arm}
        \label{fig:p4}
    \end{subfigure}
    \hfill
    \begin{subfigure}{0.49\textwidth}
        \vspace{4pt}
        \centering
        \includegraphics[width=8.4cm]{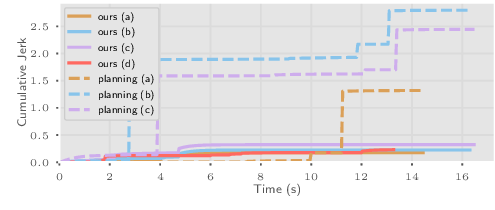}
        \vspace{-4pt}
        \caption{
            Cumulative End-effector Jerk
        }
        \label{fig:jerk}
    \end{subfigure}

    \caption{The Frankie mobile manipulator moving from a starting pose to a desired pose using our proposed motion controller where robot opacity increases with time. (a) Goal is directly in front of the robot. (b) Goal is to the side of the robot. (c) Goal is directly behind the robot. 
    (d) The goal is moving for 12\unit{s} (red line) and the end-effector (blue line) catches and tracks it.
    (e) The manipulability of the arm throughout (a)-(d). (f) Cumulative end-effector jerk comparing the holistic and planning based approaches.} \label{fig:p0}
    \vspace{-8pt}
\end{figure*}

\subsection{Experiment 3: Repetitive Pick and Place Task}

We evaluate the merits of our approach in a real-world repetitive pick-and-place task. In this experiment, the mobile manipulator must pick common household objects from a container and place them on a table located 3\unit{m} across the room from the original container.

To make our system taskable and responsive to many competing goals and states, we use behaviour trees. A behaviour tree is a directed acyclic graph with a single root vertex. Vertices represent behaviours and edges represent the flow of control. Vertex execution is governed by ticks which are an activation signal. The ticks are propagated through the tree at a defined frequency. Upon receiving a tick, the vertex returns running if it is still executing, success if its goal was achieved, or otherwise failure. This process makes behaviour trees highly reactive, as the whole tree gets processed on every tick. A comprehensive description of behaviour trees can be found in \cite{bt1}.

We have implemented a behaviour tree to complete this pick-and-place task. We assume the robot is localised within a two-dimensional map of the environment, and the container/table locations within this map are known. We use GG-CNN to perform grasp point synthesis \cite{doug} and a RealSense D435 eye-in-hand camera to provide depth images. GG-CNN consumes depth images and outputs the best grasp pose visible from the depth image.

The behaviour tree will first run a branch which recovers the arm from any errors (encountered by joint position, velocity, force limits etc.) and moves the arm into a ready pose and the base to the initial pose. Any unhandled failure will cause the tree to return to this branch where errors are recovered and the platform is re-initialised. The second branch of the tree will repeat on success. This means that the pick and place task will repeat unless there is a failure.

In the second branch, the robot will first attempt to grasp an object. The grippers are opened if required, and then the platform moves to a defined pickup pose using our Holistic Motion Controller from which the container holding the objects is visible to the eye-in-hand camera. Subsequently, a grasp is attempted. In a grasp attempt, the Holistic Motion Controller guides the robot towards the grasp pose output from GG-CNN. When the distance to the desired grasp pose is greater than 0.25\unit{m}, the output from GG-CNN is continuously updating the Holistic Motion Controller goal. Due to the minimum sensing distance of the RealSense D-435 camera, the grasp pose is locked in when it is 0.25\unit{m} away from the camera. After a grasp attempt, the arm retreats to a ready configuration where grasp success is detected. A failure is detected by the gripper being fully closed (i.e. the hand is empty), or the arm encounters an error. If a failure is detected, the grasp attempt is repeated.

Upon a successful grasp, the platform moves to a dropoff pose where the object is placed. The pickup and dropoff locations are separated by approximately 3\unit{m}. The pick and place sequence is then repeated.

To perform a comparison to prior work, we repeated the pick and place task while using motion planners instead of our holistic controller. In this implementation, we use the internal Omron planner to control the mobile base, and the RRTConnect motion planner from OMPL to control the arm \cite{ompl}. This mimics the robot control approach from \cite{mm0}. The behaviour tree and vision stack remain the same, except we only use the first output from GG-CNN due to the open-loop nature of motion planners.

\section{Results} \label{sec:res}

The average execution time of the holistic motion controller during the experiments was $4.8$\unit{ms} using an Intel i5 NUC with 4 cores at 2.3GHz in a single threaded process -- a control rate just above $200$\unit{Hz}. We use the following values for the controller parameters: $\eta = 1$, $\rho_i = 50\deg$, and $\rho_s = 2\deg$ in (\ref{eq:gain2}), $k_\epsilon = 0.5$ in (\ref{eq:kappa}), $k_a = 0.01$ in (\ref{eq:ka}), and $\beta = 1$ in (\ref{eq:pbs}). The effects of changing the controller parameters is discussed in Experiment 2 results. In practice we found that performance is sound within $50\%$ to $200\%$ of the chosen gain values.

\subsection{Experiment 1 Results}
We display the robot motion of Frankie from Experiment 1 in Figure \ref{fig:p0} using the Swift Simulator \cite{rtb}, however we refer readers to the supplementary video which shows the robot motion with a comparison to the non-holistic planning-based approach, and the motion of the holonomic Omni-Frankie robot. We also provide code examples of our motion controller running on the Frankie and Omni-Frankie mobile manipulators at \href{https://jhavl.github.io/holistic}{jhavl.github.io/holistic}.

As shown by Figure \ref{fig:p0}, the mobile manipulator achieves all four poses set out by the experiment. These motions took 5.42, 6.17, 6.17, and 13.12 seconds to complete and correspond to Figure \ref{fig:p1}, \ref{fig:p2}, \ref{fig:p3}, and \ref{fig:p4} respectively. We repeated Experiment 1 using a method representative of the non-holistic planning-based approach from \cite{mm0}. Using this approach, the robot took 14.32, 16.17, and 16.45 seconds to complete the motion respectively which is between two and three times slower than the holistic reactive approach. Note that there is no planning-based approach for Experiment 1d) where the goal moves, due to the open-loop nature of planners.

Figure \ref{fig:p1} shows the simplest of the motions, and the base and arm cooperate so the arm  remains well-conditioned and not overly stretched out. Figure \ref{fig:p2} shows a slightly more complicated motion as the base must rotate and translate in cooperation with the arm to reach the goal. The base rotates in cooperation with the arm motion preventing the arm from approaching its joint limits. Figure \ref{fig:p3} shows a very complex task requiring close cooperation between base and arm. The base rotates and translates to ensure the arm does not encounter joint limits in the motion.  Figure \ref{fig:move} shows that our controller can instantaneously, and smoothly react to a changing goal pose. Notably, these motions are not the result of a computed trajectory, it is an emergent behaviour of the reactive holistic motion controller. Figure \ref{fig:p4} shows that in all four motions, the robot arm remains well-conditioned throughout when using our proposed controller.

Figure \ref{fig:jerk} shows the cumulative end-effector jerk in each motion when using our proposed holistic controller, and the non-holistic planning based approach where the jerk from the initial robot motion has been ignored. Lower cumulative jerk indicates smoother overall motion and we see that the holistic approach generates quantitatively smoother motion than previous work which involves disjointed stop-start motion.

\subsection{Experiment 2 Results}

Table \ref{tab:comp} shows that for both holonomic and non-holonomic mobile manipulators, our holistic controller outperforms the non-holistic motion planners. Failures in the holistic controller are caused by local minima, while motion planning failures are the result of a solution not being found within 5\unit{s}.

Table \ref{tab:kappa} shows the effect of $k_\epsilon$ (\ref{eq:kappa}) on performance. To maximise the ability for the robot to complete follow-on tasks, including grasping, it is important to minimise the base to end-effector angle $\theta_\epsilon$. Making $k_\epsilon$ large will minimise $\theta_\epsilon$ but leads to more controller failures due to ill-conditioned arm configurations leading to local minima. Table \ref{tab:kappa} shows that  $k_\epsilon = 0.1 \sim 0.5$ provides suitable results.

Table \ref{tab:manip} shows the effect which $\emat{J}_m$ has in (\ref{eq:jmref}) on performance. These results show that low arm manipulability $m(\vec{q}_a)$ leads to operational failures which is consistent with \cite{neo}. When manipulability is not optimised with $\emat{J}_m = (\mat{0}_b, \ \mat{0}_a)$ and when manipulability of the whole platform is optimised with $\mat{J}^{ba}_m(\vec{q})$ the average final arm manipulability is low which leads to increased failures. Therefore, it is important to maximise the arm's ability to exploit its limited workspace.

\subsection{Experiment 3 Results}

We repeated Experiment 3 ten times for a total of 100 objects picked and placed. The robot successfully picked and placed all 100 objects with 113 grasp attempts. We obtained the same grasp performance when we repeated the task while using the motion planning approach instead of our holistic controller. In this experiment, there were no failures caused by the motion control or robot actuation. Minor failures were caused by the vision pipeline leading to incorrect grasp points from the GG-CNN network. Failed grasp attempts were caught by the behaviour tree and caused the grasp to be re-attempted. A full run of Experiment 3 is shown in our supplementary video.

\newcolumntype{M}[1]{>{\centering\arraybackslash}m{#1}}

\begin{table}[t!]
    \centering
    \renewcommand{\arraystretch}{1.3}
    \vspace{4pt}
    \caption{Experiment 2a) Comparing Number of Failures over 1000 Random Reaching Tasks}
    \label{tab:comp}
    \begin{tabular}{  M{1.6cm} | M{1.6cm} | M{1.6cm} | M{1.6cm} }
    \hline
    \multicolumn{2}{c|}{Frankie (non-holonomic)} & \multicolumn{2}{c}{Omni-Frankie (holonomic)} \\
    \hline
    Motion Planning & Holistic (ours) & Motion Planning & Holistic (ours)\\
    \hline\hline
    71 & 35 & 71 & 36\\
    \hline
    \end{tabular}
\end{table}

\begin{table}[t!]
    \centering
    \renewcommand{\arraystretch}{1.3}
    \caption{Experiment 2b) Effect of $k_\epsilon$ on Performance over 1000 Random Reaching Tasks}
    \label{tab:kappa}
    \begin{tabular}{  M{1.7cm} | M{1.1cm} | M{1.1cm} | M{1.1cm} | M{1.1cm} }
    \hline
    $k_\epsilon$ & 0.0 & 0.1 & 0.5 & 1.0 \\
    \hline\hline
    Failures & 38 & 34 & 39 & 76 \\
    Mean Final $\theta_\epsilon$ & $22.91\deg$ & $10.31\deg$ & $4.58\deg$ & $3.43\deg$ \\
    \hline
    \end{tabular}
\end{table}

\begin{table}[t!]
    \centering
    \renewcommand{\arraystretch}{1.3}
    \caption{Experiment 2c) Effect of $\emat{J}_m$ on Performance over 1000 Random Reaching Tasks}
    \label{tab:manip}
    \begin{tabular}{  M{2.2cm} | M{1.2cm} | M{1.2cm} | M{1.8cm} }
    \hline
    $\emat{J}_m$ & $(\mat{0}_b, \ \mat{0}_a)$ & $\mat{J}^{ba}_m(\vec{q})$ & $(\mat{0}_b, \ \mat{J}^a_m(\vec{q}_a))$ \\
    \hline\hline
    Failures & 161 & 161 & 39  \\
    Mean Final $m(\vec{q}_a)$ & 0.032 & 0.033 & 0.079  \\
    \hline
    \end{tabular}
\end{table}

For all robotic applications, task performance time is an important performance metric.  While we measure and report our motion times in Table \ref{tab:results}, we were not able to find timing results in other papers reporting similar robotic tasks. For example, in \cite{mm0} the robot picks an object from a bench-top containing several household objects, while in \cite{mm9}, the robot picks an object from a bench-top containing a single household object. In \cite{mm7}, a mobile manipulator must pick and place cups from a bench where no base motion is required. The task and movement requirements are identical to our task, or easier in the case of \cite{mm7}, so it is interesting to compare motion times for various approaches. It would be preferable to use published timing results but, in its absence, we have reverse engineered these from analysis of published supplementary video material and it is summarized in Table \ref{tab:results}.
We could encourage future authors to include time and distance data in their papers to encourage meaningful comparisons.

The timing results from Experiment 3, using motion planners, is also shown in Table \ref{tab:results} under the \emph{planning} approach. Grasp time is defined as the time spent in the grasp motion, which results in the object being grasped (including repeat grasp attempts where applicable), and pick and place time is defined as grasp time plus travelling to the drop-off point and placing the object. The planning based approaches all offered good grasp success rates but took much longer to complete the task. As shown in Table \ref{tab:results}, our mean grasp time is 3 times faster than the motion planning approach and \cite{mm0} while being 6 times faster than \cite{mm9}. Our average pick and place time is 2 times faster than the motion planning approach and 3 times faster than \cite{mm7}. These timings are consistent with the timing comparison in Experiment 1. Results from these experiments suggest that our holistic and reactive motion controller can significantly improve task execution time on a mobile manipulator while also being reliable and robust to task completion. The resulting motion is graceful: intuitive, smooth and fast \cite{grace}, while the time spent with the robot stationary has been completely eliminated. 

\begin{table}[t!]
    \centering
    \renewcommand{\arraystretch}{1.3}
    \vspace{6pt}
    \caption{Pick and Place Timing Comparison}
    \label{tab:results}
    \begin{tabular}{  M{1.4cm} | M{2.3cm} | M{3.3cm} }
    \hline
    Approach & Mean Grasp Time & Mean Pick and Place Time \\
    \hline\hline
    ours & 5.2\unit{s} & 16.8\unit{s} \\
    planning & 15.6\unit{s} & 35.2\unit{s} \\
    \cite{mm0} & 15.4\unit{s} & - \\
    \cite{mm9} & 30\unit{s} & - \\
    \cite{mm7} & - & 51\unit{s} \\
    \hline
    \end{tabular}
\end{table}

\section{Conclusions}

In this paper, we proposed a novel motion controller which provides reactive and holistic control of both non-holonomic and holonomic mobile manipulation platforms. Our experiments demonstrate that our purely reactive and holistic approach performs faster, more efficiently, and more robustly than related work. In combination with a behaviour tree, our reactive system becomes taskable and capable of error recovery as demonstrated by a mobile pick and place task where the motion is smooth, intuitive and fast.

Thanks to the versatility of behaviour trees, our behaviour tree can be readily modified to perform different tasks, or extended with \cite{bt0} to automatically generate behaviour trees based on natural language input. In the future, we would like to implement perception related behaviours, such as object recognition and collision detection, which makes a wider range of tasks solvable. Additionally, with more perception we can incorporate the reactive collision avoidance from \cite{neo} into the holistic motion controller.

We have provided full examples of our motion controller running on the Frankie and Omni-Frankie mobile manipulators at \href{https://jhavl.github.io/holistic}{jhavl.github.io/holistic}.

\vspace{8pt}
\bibliographystyle{IEEEtran} 
\bibliography{ref}

\end{document}